\documentclass[11pt,a4paper]{article}

\usepackage[utf8]{inputenc}
\usepackage[english]{babel}
\usepackage{amsmath,amssymb}
\usepackage{graphicx}
\usepackage{booktabs}
\usepackage{array}
\usepackage{subfigure}
\usepackage{algorithm}
\usepackage{algpseudocode}
\usepackage{multirow}
\usepackage{cite}     
\usepackage{cleveref} 
\usepackage{xcolor}
\usepackage{siunitx}
\usepackage{tikz}
\usepackage[top=2.5cm, bottom=2.5cm, left=2.5cm, right=2.5cm]{geometry}

\newcolumntype{L}[1]{>{\raggedright\let\newline\\\arraybackslash\hspace{0pt}}p{#1}}
\newcolumntype{C}[1]{>{\centering\let\newline\\\arraybackslash\hspace{0pt}}p{#1}}
\newcolumntype{R}[1]{>{\raggedleft\let\newline\\\arraybackslash\hspace{0pt}}p{#1}}

\newcommand{\al}{\alpha}

\newcommand{\de}{\delta}
\newcommand{\ep}{\varepsilon}

\newcommand{\dde}{\dot{\de}}
\newcommand{\dpsi}{\dot{\psi}}

\newcommand{\bx}{\boldsymbol{x}}     

\newcommand{\bz}{\boldsymbol{z}}

\newcommand{\bP}{\boldsymbol{P}}
\newcommand{\bQ}{\boldsymbol{Q}}

\newcommand{\bzero}{\boldsymbol{0}}

\newcommand{\dbx}{\dot{\bx}}     

\newcommand{\dr}{\dot{r}}

\newcommand{\dv}{\dot{v}}

\newcommand{\dx}{\dot{x}}
\newcommand{\dy}{\dot{y}}

\newcommand{\bbR}{\mathbb{R}}    

\newcommand{\calB}{\mathcal{B}}    

\newcommand{\tr}{\operatorname{tr}}  

\providecommand{\norm}[1]{\lVert#1\rVert}  

\newcommand{\dV}{\dot{V}}

\DeclareMathOperator{\vol}{vol}
\newcommand{\SOS}[1]{\Sigma[#1]}
\newcommand{\SOSs}[1]{\Sigma^{+}[#1]}

\newcommand{\yG}{y_G}
\newcommand{\dyG}{\dy_G}
\newcommand{\deone}{\de_1}
\newcommand{\detwo}{\de_2}
\newcommand{\ddeone}{\dot{\deone}}
\newcommand{\ddetwo}{\dot{\detwo}}
\newcommand{\ddde}{\ddot{\de}}
\newcommand{\dddde}{\dddot{\de}}

\newcommand{\ed}{\dot{e}}

\newcommand{\Lprev}{L_P}
\newcommand{\Tprev}{T_P}

\newcommand{\Fyf}{F_{y,f}}
\newcommand{\Fyr}{F_{y,r}}

\newcommand{\alf}{\al_f}
\newcommand{\alr}{\al_r}
\newcommand{\ali}{\al_i}
\newcommand{\alflim}{\bar{\al}_f} 
\newcommand{\alrlim}{\bar{\al}_r} 
\newcommand{\alilim}{\bar{\al}_i}

\newcommand{\Phif}{\Phi_f}
\newcommand{\Phir}{\Phi_r}
\newcommand{\Phii}{\Phi_i}

\newcommand{\etai}{\eta_i}

\newcommand{\Vstep}[1]{V_{#1}}
\newcommand{\Pstep}[1]{\bP_{#1}}
\newcommand{\lambdastep}[1]{\lambda_{#1}}
\newcommand{\mustep}[1]{\mu_{#1}}
\newcommand{\dVlinstep}[1]{\dV_{#1,\text{lin}}}
\newcommand{\rstep}[1]{\rho_{#1}}

\title{Global stability of vehicle-with-driver dynamics via Sum-of-Squares programming}

\author{%
  Martino Gulisano\thanks{Email: \texttt{martino.gulisano@phd.unipi.it}}\hspace{0.5em} and\hspace{0.5em}
  Marco Gabiccini\thanks{Email: \texttt{marco.gabiccini@unipi.it}}\\[0.5em]
  \small\textsc{Università di Pisa, Dipartimento di Ingegneria Civile e Industriale}\\
  \small\textsc{Largo L. Lazzarino 1, 56122 Pisa, Italy}
}

\date{}

\begin{document}

\maketitle


\begin{abstract}
This work estimates safe invariant subsets of the Region of Attraction (ROA) for a seven-state vehicle-with-driver system, capturing both asymptotic stability and the influence of state-safety bounds along the system trajectory. Safe sets are computed by optimizing Lyapunov functions through an original iterative Sum-of-Squares (SOS) procedure.

The method is first demonstrated on a two-state benchmark, where it accurately recovers a prescribed safe region as the 1-level set of a polynomial Lyapunov function. We then describe the distinguishing characteristics of the studied vehicle-with-driver system: the control dynamics mimic human driver behavior through a delayed preview-tracking model that, with suitable parameter choices, can also emulate digital controllers. To enable SOS optimization, a polynomial approximation of the nonlinear vehicle model is derived, together with its operating-envelope constraints.

The framework is then applied to understeering and oversteering scenarios, and the estimated safe sets are compared with reference boundaries obtained from exhaustive simulations. The results show that SOS techniques can efficiently deliver Lyapunov-defined safe regions, supporting their potential use for real-time safety assessment, for example as a supervisory layer for active vehicle control.
\end{abstract}

\section{Introduction}

Vehicle stability is a cornerstone of road safety because any departure from stable motion can quickly escalate into accidents and injuries.
Engineers have therefore devoted substantial effort to early detection of the onset of instability and to designing supervisory layers---such as ESC, active yaw-rate or torque-vectoring controllers, and coordinated brake- or steer-by-wire actuation---that prevent dangerous operating conditions.
Even advanced supervisory strategies still depend on empirically tuned thresholds and controller parameters, often lacking actual stability certificates grounded on mathematical guarantees.
The role of the driver further complicates this picture: its influence on the coupled dynamics is frequently ignored, even though driver actions can trigger instability \cite{Mastinu:GlobalStabilityRoad:2023}.
Developing rigorous model-based stability certificates that delineate safe operating regions across driving maneuvers, driver behaviors, and hardware configurations would help close this gap while directly informing controller design.
These needs motivate the adoption of advanced estimation techniques for safe operative regions endowed with provable guarantees.

\subsection{Literature review}
In the scientific literature there are several approaches to vehicle stability.
Many application-oriented studies adopt empirical limits on vehicle state variables, their thresholds being chosen experimentally or based on engineering practice.
It is quite common to impose tunable bounds on the vehicle sideslip angle to stabilize the vehicle, for example in the design of ESC systems \cite{Chen:ESC:2014} or torque vectoring controllers \cite{Lenzo:YawRateSideslip:2021}.
Alternatively, some methods avoid relying on the vehicle sideslip angle, whose direct measurement is challenging and typically requires estimation.
For instance, in \cite{Daher:ggdiagram:2017} the authors proposed a vehicle stability controller that sets bounds on the vehicle longitudinal and lateral accelerations, i.e., on the g-g diagram.

On the other hand, a large body of work takes a more theoretical approach by relying on simplified mathematical models of the vehicle dynamics---typically single-track or double-track models that describe the lateral and yaw motion of the vehicle---and analyzing their stability as dynamical systems.
It is standard textbook practice \cite{Guiggiani:ScienceVehicleDynamics:2023,Mastinu:RoadOffRoadVehicle:2014} to assume some constant inputs (e.g., fixed longitudinal speed and steering angle) and to linearize the vehicle dynamics around the corresponding steady-state motion, i.e., straight-ahead or steady-cornering conditions.
The eigenvalues analysis of the linearized system allows to assess the stability of the corresponding equilibrium point.

Because tire characteristics introduce nonlinearities in the vehicle dynamics, multiple equilibrium points may coexist.
Early contributions in this area by Pacejka \cite{Pacejka:TireVehicleDynamics:2012} already showed that a two-state single-track model with nonlinear axle characteristics may exhibit three distinct steady-state points.
The study of how the number and stability of these equilibria change as a function of vehicle parameters is known as bifurcation analysis \cite{Strogatz:NonlinearDynamicsChaos:2019}.
Comprehensive analyses using continuation methods have identified various bifurcation phenomena---including saddle-node, Hopf, homoclinic, heteroclinic, and Bogdanov-Takens bifurcations---by varying driver inputs and tire characteristics \cite{DellaRossa:bifurcation:2012}.
Complementary approaches combined phase-plane and handling diagram methods to investigate how vehicle parameters such as roll stiffness and center of mass position can also induce bifurcation \cite{Farroni:CombinedUsePhase:2013}, and to compare the stability performance of different steering control architectures \cite{Lai:bifurcation:2021}.

For a stable and desirable steady-state motion, a crucial question concerns the size of its Region of Attraction (ROA), i.e, the set of initial conditions from which trajectories converge asymptotically to the equilibrium.
Estimating the ROA allows to quantify how large deviations from the steady-state can be tolerated before the vehicle loses stability.
Early efforts to compute the ROA analytically applied Lyapunov's direct method with quadratic Lyapunov functions to single-track models with cubic polynomial axle characteristics \cite{Johnson:NonlinearLateralStability:1984}.
More recent study have conducted ROA estimation with more complex tire models. Linear parameter-varying (LPV) formulations with LuGre tire dynamics have been used to derive Lyapunov-based stability conditions \cite{Hashemi:stability:2016}. Alternative methods based on local linearization at multiple operating points have also been proposed \cite{Huang:StabilityRegionsVehicle:2020}. Other works have focused on quantitative stability indicators, such as Lyapunov exponents to measure convergence rates \cite{Sadri:LyapunovExponents:2013} or energy dissipation as a measure of stability \cite{Meng:Dissipation:2022}.

For systems with polynomial dynamics, Lyapunov-based ROA estimation can be performed efficiently using Sum-of-Squares (SOS) programming techniques \cite{Jarvis-Wloszek:ControlApplicationsSOS:2005,Topcu:SOS:2008}. 
This approach exploits the fact that a sufficient condition for polynomial Lyapunov candidates to be positive-definite can be imposed by requiring them to be written as SOS polynomials. 
This SOS requirement can be formulated as a convex constraint in semidefinite programs (SDPs), enabling systematic computation of polynomial Lyapunov functions. Several applications to vehicle lateral dynamics have already used SOS techniques to estimate stability regions \cite{ImaniMasouleh:RegionAttractionAnalysis:2018,Tamba:NonlinearStabilityAnalysis:2018,Ribeiro:SOSapproach:2020,Ribeiro:statefeedbackSOS:2022,Zhu:SOSBasedVehicle:2022}.

\subsection{Paper's contributions and organization}
In the present work we extend the application of SOS programming to estimate the safety sets for a vehicle-with-driver system. While SOS techniques have already been successfully applied to vehicle lateral dynamics \cite{ImaniMasouleh:RegionAttractionAnalysis:2018,Tamba:NonlinearStabilityAnalysis:2018,Ribeiro:SOSapproach:2020,Ribeiro:statefeedbackSOS:2022,Zhu:SOSBasedVehicle:2022}, our contribution focuses on two important aspects. 
First, we explicitly account for the driver's action, performing a \emph{global stability analysis} of the coupled vehicle-with-driver system. Recent works \cite{Mastinu:GlobalStabilityRoad:2023,Mastinu:HowDriversLose:2024} have shown that the driver can be a source of instability for the overall system. Although human driver modeling is a challenging task, simplified models available in the literature \cite{Plochl:DriverModels:2007} allow for preliminary estimates. While certainly not exact, these models enable qualitative reasoning and provide results that are also quantitatively consistent with experimental observations \cite{Mastinu:HowDriversLose:2024}.
Second, we explore how to constrain the Lyapunov function search to identify \emph{safe} invariant subsets of the ROA. Unlike traditional SOS approaches that focus on asymptotic convergence to equilibrium, we require that trajectories remain within prescribed state boundaries throughout their entire time evolution.
In our setting, constraining the trajectories to these safe invariant subsets guarantees that the polynomial model used in the SOS formulation remains valid and thus provides an accurate local approximation of the original nonlinear dynamics.

The remainder of the paper is organized as follows.
Sec.~\ref{sec:roa} introduces the SOS framework on a two-state benchmark, detailing each step of the iterative Lyapunov search and illustrating how the resulting level sets approximate a prescribed safe region.
Sec.~\ref{sec:vehicle} presents the seven-state vehicle-with-driver system, specifies the polynomial approximations required to apply the SOS framework, and highlights the state constraints needed to preserve fidelity to the original non-polynomial model.
Sec.~\ref{sec:results} reports two case studies—understeering and oversteering scenarios—examining the intersections of the estimated safe sets with relevant slices of the state space and comparing the SOS-certified regions with simulation-based target envelopes.
Finally, Sec.~\ref{sec:conclusion} summarizes the main findings and outlines directions for future research.

\section{ROA estimation via Lyapunov functions}
\label{sec:roa}

\subsection{General formulation}
\label{subsec:genform}
\begin{figure}
    \centering
    \subfigure[First iteration of the SOS-iterative procedure described in Sec.~\ref{subsec:iterSOS}: $\Omega=\{\bx:V(\bx)\le\rho\}$ (pink region, red boundary), elliptical shape set (blue region), $\{x:\dV(\bx)>0\}$ (purple region). Unstable limit cycle in solid black line.]{%
        \includegraphics[width=0.48\linewidth]{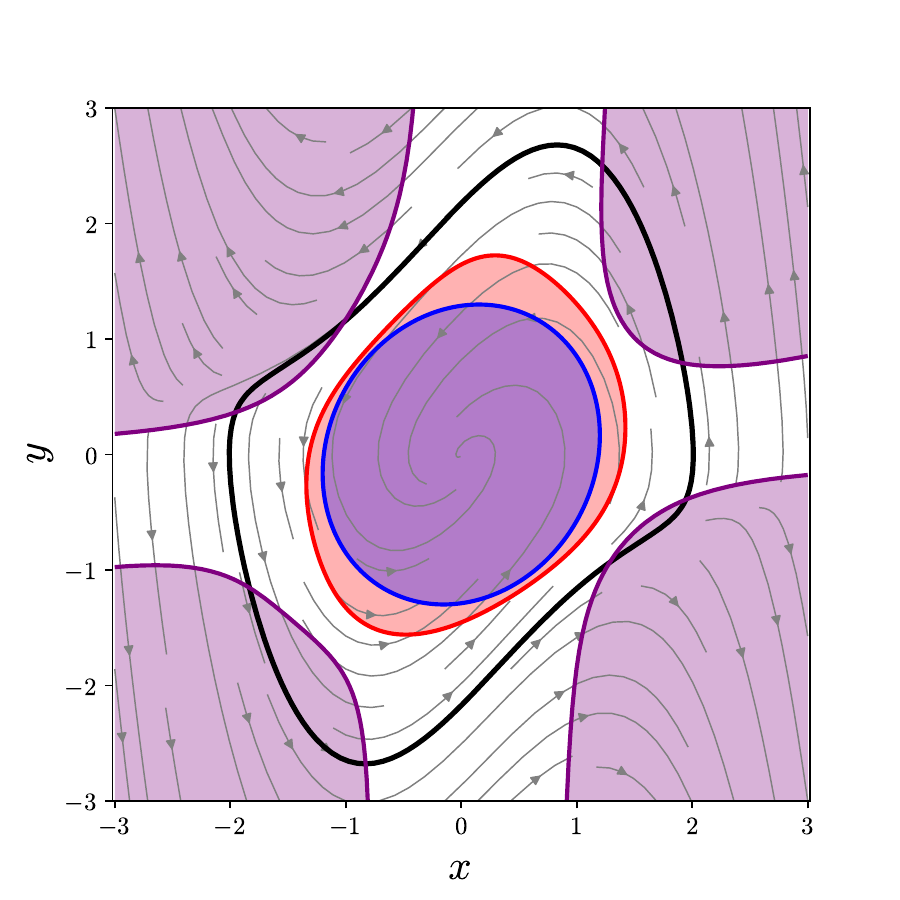}
        \label{fig:vdp:iter}}
    \hfill
    \subfigure[Final ROA estimate (pink region) with ROA boundary (red line) lying almost on top of the limit cycle (solid black line). Increasingly tight approximation of the ROA boundaries during iterations are represented as brown lines.]{%
        \includegraphics[width=0.48\linewidth]{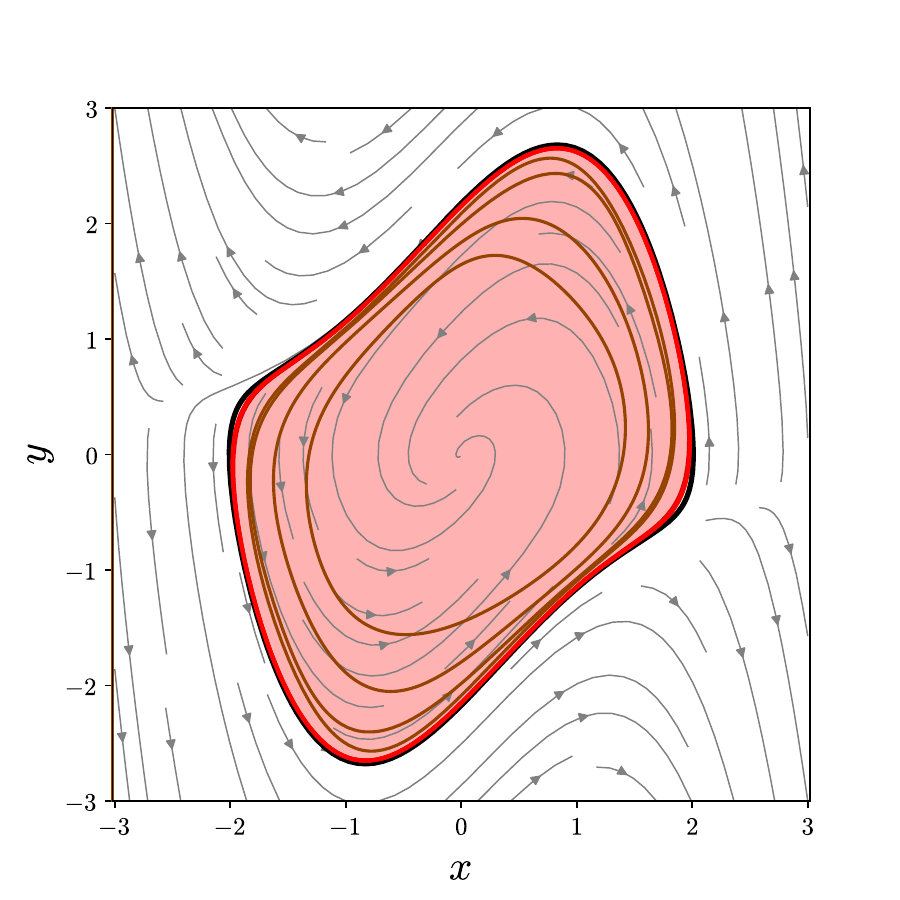}
        \label{fig:vdp:ROAiter}}
    \caption{Phase portraits of time-reversed Van der Pol Oscillator with ROA estimates.}
    \label{fig:vdp}
\end{figure}

We first illustrate the ROA estimation procedure on a two-state polynomial system, i.e., the \emph{time-reversed} Van der Pol oscillator, whose dynamics $\dbx = f(\bx)$ is given by
\begin{equation}
    \label{eq:vdp}
    \begin{cases}
        \dx = -y \\
        \dy = x + \mu \, (x^2-1) y 
    \end{cases}
\end{equation}
The system has an asymptotically stable equilibrium at the origin and an unstable (saddle-type) limit cycle that marks the boundary of the ROA for the origin. The distortion of this region increases with greater values of $\mu$: here we pick $\mu=1$, since it already provides a challenging ROA shape to approximate. The phase portrait of system~\eqref{eq:vdp} is illustrated in Figure~\ref{fig:vdp:iter} along with the ustable limit cycle (black curve). 

A Lyapunov function $V(\bx)$ is a differentiable energy-like function whose value decreases along the system trajectories. As is well known~\cite{SlotineLi:book:1991}, \emph{if one can find} a function $V(\bx)$ satisfying (i) $V(\bx)\succ0$ and (ii) $\dV(\bx)=\nabla V(\bx) f(\bx)\prec0$, then the origin is asymptotically stable. This is the case for the \emph{time-reversed} Van der Pol oscillator. Here, $V(\bx)\succ0$ means that $V$ is positive definite---i.e., $V(\bzero)=0$ and $V(\bx)>0$ in a neighborhood $\calB_0$---while $\dV(\bx) \prec 0$ indicates negative definiteness. 

Moreover, \emph{whenever} such a Lyapunov function exists, any sublevel set
$\Omega=\{\bx:V(\bx)\le\rho\}$ in which $\dV(\bx)<0$ for all $\bx \neq \bzero$, by the Local Invariant Set theorem~\cite[p.69]{SlotineLi:book:1991}, constitutes and inner estimate of the ROA, since it
guarantees that all the trajectories starting within $\Omega$ converge to the origin---as it is the largest invariant set in $\Omega$ for system~\eqref{eq:vdp}. 

The brown curves in Figure~\ref{fig:vdp:ROAiter} depict increasingly accurate approximations of the ROA boundary for system \eqref{eq:vdp}. At convergence, the red curve delineates the boundary of the final ROA estimate—the entire pink region—and it lies essentially on top of the true limit cycle.

The problem of estimating the ROA can be cast as the search for the \emph{largest sublevel set} $\Omega$, i.e., with highest volume. In particular, one can seek a function $V(\bx)$, a level $\rho$, and a shape set $S$ and pose the problem as follows:
\begin{subequations}\label{eq:genROA}
    \begin{align}
        \underset{V(\bx), \rho, S}{\text{maximize}} \quad & \vol (S) \label{eq:genROAcost} \\
        \text{s.t.} \quad           & V(\bx)\succ0                  \label{eq:genROAv}  \\
        \phantom{\text{s.t.} \quad} & \dV(\bx)\prec0                \label{eq:genROAvdot}\\
        \phantom{\text{s.t.} \quad} & \Omega \subseteq \{x:\dV(\bx)\le0\} \label{eq:genROAin}\\
        \phantom{\text{s.t.} \quad} & S \subseteq \Omega. \label{eq:genROAshape}
    \end{align}
\end{subequations}
Here, $S$ denotes the shape set (typically a non-degenerate ellipsoid $S=\{\bx: \bx^T \bP \bx \le 1\}$) whose volume can be computed analytically. Introducing $S$ this way provides an explicit surrogate objective $\vol (S)$ for $\vol (\Omega)$, allowing to systematically seek the largest certified inner approximation of the ROA by the chain of inclusions~\eqref{eq:genROAin}--\eqref{eq:genROAshape}. 
This inclusion chain is illustrated for system \eqref{eq:vdp} in Fig.~\ref{fig:vdp:iter}, where 
$S$ is the blue ellipse, $\Omega$ is the pink region (red boundary), and the set $\{\bx:\dV(\bx)>0\}$ is highlighted in purple (showing the complement of $\dV(\bx)\le0$ for readability).

The problem \eqref{eq:genROA} is an optimization over a function space, and is intractable in its general form. 
To obtain a computationally feasible formulation, we rely on Sum-of-Squares (SOS) programming~\cite{Jarvis-Wloszek:ControlApplicationsSOS:2005}. 
The key idea is to restrict the search for $V(\bx)$ to SOS polynomials and reformulate~\eqref{eq:genROAv}--\eqref{eq:genROAshape} 
as SOS conditions. 

\subsection{SOS formulation}
\label{subsec:SOSform}
A polynomial $p(\bx)$ is SOS if it can be written as $p(\bx) = \sum_i q_i^2(\bx)$ for some polynomials $q_i(\bx)$. 
We denote by $\SOS{\bx}$ the space of SOS polynomials in $\bx$. 
Note that $p(\bx)\in\SOS{\bx}$ implies that $p(\bx)$ is a positive polynomial (a.k.a. nonnegative), i.e., $p(\bx) \geq 0$ for all $\bx$\footnote{The converse does not hold, as exemplified by the Motzkin polynomial~\cite{Motzkin:paper:1967}, which is globally nonnegative yet admits no representation as a sum of squares.}.
Checking (or enforcing) that a polynomial $p(\bx)$ is SOS, i.e. $p(\bx)\in\SOS{\bx}$, amounts to solving a semidefinite program (SDP) whose decision variables are linked to the coefficients of $p(\bx)$ through its Gram-matrix representation. 

Specifically, any polynomial $p(\bx)$ can be decomposed as $p(\bx) = \bz(\bx)^T \bQ \bz(\bx)$, where $\bz(\bx)$ is the vector of monomials and $\bQ=\bQ^T$ is the Gram matrix. As an example, a fourth-degree polynomial in two variables $x$ and $y$ can be written as 
\begin{align}
p(x,y) = \sum_{i=1}^{6}\sum_{j}^{6} Q_{ij} z_i(x,y) z_j(x,y),
\end{align}
where $\bz(x,y)=[1\;\; x\;\; y\;\;  x^2\;\; xy\;\; y^2]^T$, $z_i(x,y)$ is the $i-$th element of $\bz(x,y)$, and $Q_{ij}$ the $(i,j)-$th element of $\bQ\in\bbR^{6\times 6}$.
In general, for a $2r$-degree polynomial in $n$ variables, $\bz(\bx)$ has $m=\binom{n+r}{n}$ monomials and $\bQ\in\bbR^{m\times m}$.
Checking (or enforcing) $p(\bx)\in\SOS{\bx}$ reduces to solving an SDP with $\frac{m(m+1)}{2}$ decision variables---the independent entries of the symmetric matrix $\bQ$---subject to the linear matrix inequality $\bQ \succeq 0$, which is a convex conic constraint.

Moreover, to certify that a polynomial $p(\bx)$ is \emph{strictly positive definite} we can subtract a small positive term $\ep\norm{\bx}^2$, with $\ep>0$, and require that the auxiliary polynomial $a(\bx) = p(\bx)-\ep\norm{\bx}^2$ belongs to $\SOS{\bx}$. Denoting by $\SOSs{\bx}$ the subset of strictly-positive SOS polynomials, the existence of such an $\ep$ then suffices to conclude that $p(\bx)\in\SOSs{\bx}$.

The problem~\eqref{eq:genROA} can be recast as
\begin{subequations}\label{eq:SOSROA}
    \begin{align}
        \underset{V(\bx), \rho, \bP, \lambda(\bx), \mu(\bx)}{\text{minimize}} \quad & \tr(\bP) \label{eq:SOSROAcost} \\
        \text{s.t.} \quad           & V(\bx) \in \SOSs{\bx}     \label{eq:SOSROAv}  \\
        \phantom{\text{s.t.} \quad} & V(\bzero) = 0           \label{eq:SOSROAv0}  \\
        \phantom{\text{s.t.} \quad} & \norm{\bx}^{2d} (V(\bx)-\rho)+\lambda(\bx) \dV(\bx) \in \SOS{\bx} \label{eq:SOSROAin} \\
        \phantom{\text{s.t.} \quad} & \norm{\bx}^{2d_1} (\bx^T \bP \bx -1) + \norm{\bx}^{2d_2} \mu(\bx) (V(\bx)-\rho) \in \SOS{\bx} \label{eq:SOSROAshape}
    \end{align}
\end{subequations}
First, we notice that~\eqref{eq:SOSROA} involves an ellipsoidal shape set. 
The volume of an ellipsoid in the form is $\bx^T \bP \bx = 1$ is proportional to $\frac{1}{\sqrt{\det(\bP)}}$. To inflate the ellipsoidal shape set, in~\eqref{eq:SOSROAcost} we minimize $\tr(\bP)$, which serves as a linear proxy of $\sqrt{\det(\bP)}$ and helps formulating the problem as convex. 
Second, the positive definite constraint on $V(\bx)$ is now given by~\eqref{eq:SOSROAv} and \eqref{eq:SOSROAv0}.
Finally, the set inclusion constraints are enforced via the \emph{S-procedure}~\cite{Jarvis-Wloszek:ControlApplicationsSOS:2005}, which provides a sufficient condition through polynomial multipliers.

Consider, for instance, constraint~\eqref{eq:SOSROAin}: if there exists a multiplier $\lambda(\bx)$ such that~\eqref{eq:SOSROAin} holds, then wherever $\dV(\bx)=0$, we must have $V(\bx) \ge \rho$. 
This implies that the boundary of the region $\{\bx:\dV(\bx)\le0\}$ (in purple in Fig.~\ref{fig:vdp:iter}) lies outside $\Omega=\{\bx:V(\bx)\le\rho\}$ (red region), thus proving the inclusion~\eqref{eq:genROAin}.
The factor $\norm{\bx}^{2d}$ with $d\in\mathbb{N}^+$ ensures the validity of the S-procedure also at the origin, since $\dV(\bzero)$ is null but clearly the origin does not lie outside the ROA.

Similarly, constraint~\eqref{eq:SOSROAshape} enforces~\eqref{eq:genROAshape} via multiplier $\mu(\bx)$. 
The scaling factors $\norm{\bx}^{2d_1}$ and $\norm{\bx}^{2d_2}$ are not strictly required to guarantee the inclusion, but are introduced to balance the degree of the two addends in~\eqref{eq:SOSROAshape}. 
Without these scaling, the quadratic term $(\bx^T \bP \bx -1)$ and the higher-degree $\mu(\bx)(V(\bx)-\rho)$ often lead to ill-conditioned constraints. 
The coefficients $d_1, d_2 \in \mathbb{N}^+$ are chosen heuristically, typically between $0$ and $4$, with $d_1>d_2$.
In our experience, the proposed factors improve the feasibility of the SDP.

Unfortunately, the products $\lambda(\bx) \dV(\bx)$ in~\eqref{eq:SOSROAin} and $\mu(\bx) (V(\bx)-\rho)$ in~\eqref{eq:SOSROAshape} make the problem bilinear in the coefficients of $V(\bx)$, $\lambda(\bx)$, and $\mu(\bx)$, rendering the optimization problem non-convex. To address this issue, we propose an alternating optimization procedure, described below.

\subsection{Iterative SOS formulation}
\label{subsec:iterSOS}
There are several examples in the scientific literature of multi-step formulations for the described problem~\cite{Jarvis-Wloszek:ControlApplicationsSOS:2005,Topcu:SOS:2008}. Here we adopt a three-step iterative procedure.
The initial guess of the procedure is obtained from a feasibility SDP, in which we search for a \emph{global}%
\footnote{Only the linearized system admits such a global Lyapunov certificate with $\dV(\bx)\prec0$ for all $\bx\ne\bzero$. Subsequent SOS steps always target local Lyapunov functions, tailored to the nonlinear dynamics.}
polynomial Lyapunov for the linearized system around the origin. 
The constraints of this initialization problem are $\Vstep{0}(\bx) \in\SOSs{\bx}$ and 
$-\dVlinstep{0}(\bx)=-\nabla \Vstep{0}(\bx) \frac{\partial f(\bx)}{\partial \bx}\Big|_{\bx=\bzero} \bx \in\SOS{\bx}$. 
It is also useful to fix the scaling of the Lyapunov polynomial by constraining its value at one or more points of the state space, e.g., $\Vstep{0}(1,2)=1$.

\begin{algorithm}
    \caption{Iterative SOS procedure}
    \label{alg:sos_iter}
    \begin{algorithmic}[1]
        \State Fix Lyapunov degree $n_V$
        \State Solve the initialization SDP to obtain $\Vstep{0}$
        \For{$k = 0$ \textbf{to} $k_{\max}$}
        \Statex \hspace{\algorithmicindent}\textbf{Step 1:}
            \Repeat
                \State \textbf{try}
                \State $\big(\lambdastep{k+1}, \rstep{k+1}\big) \gets \lambda$-step$\big(\Vstep{k}, n_\lambda\big)$ \Comment{$(\lambda, \rho)$ update, Solve Problem~\eqref{eq:lam:step}}
                \If{not successful}
                    \State $n_\lambda \gets n_\lambda + 1$ \Comment{increase degree and retry}
                \EndIf
            \Until{\(\lambda\)-step succeeds or \(n_\lambda > n_\lambda^{\max}\)}
            \State $\Vstep{k} \gets \Vstep{k}/\rstep{k+1}$ \Comment{level set normalization}
            \Statex \hspace{\algorithmicindent}\textbf{End Step 1}
            \Statex %
            \Statex \hspace{\algorithmicindent}\textbf{Step 2:}
            \Repeat
                \State \textbf{try}
                \State $\mustep{k+1} \gets \mu$-step$\big(\Vstep{k}, n_\mu\big)$ \Comment{$(\mu, \bP)$ update, Solve Problem~\eqref{eq:mu:step}}
                \State \dots
            \Until{\(\mu\)-step succeeds or \(n_\mu > n_\mu^{\max}\)}
            \Statex \hspace{\algorithmicindent}\textbf{End Step 2}
            \Statex %
            \Statex \hspace{\algorithmicindent}\textbf{Step 3:}
            \State $\big(\Vstep{k+1},\Pstep{k+1}\big) \gets V\text{-step}\big(\lambdastep{k+1}, \mustep{k+1}\big)$ \Comment{$(V,\bP)$ update, Solve Problem~\eqref{eq:V:step}}
            \State break criterion on $\tr \Pstep{k+1}$ convergence 
        \State \textbf{return} $\Vstep{k+1}$
        \Statex \hspace{\algorithmicindent}\textbf{End Step 3}
        \EndFor 
    \end{algorithmic}
\end{algorithm}

A pseudocode for three-step procedure is given in Algorithm~\ref{alg:sos_iter}.
In the $\lambda$-step ({\bf Step 1}) the Lyapunov candidate $V(\bx)$ is kept fixed. We search for the largest level set satisfying~\eqref{eq:SOSROAin} by maximizing the level value $\rho$ while solving for the multiplier $\lambda(\bx)$:
\begin{subequations}\label{eq:lam:step}
    \begin{align}
        \underset{\rho, \lambda(\bx)}{\text{maximize}} \quad & \rho \label{eq:lam:obj} \\
        \text{s.t.} \quad           & \norm{\bx}^{2d} (V(\bx)-\rho)+\lambda(\bx) \dV(\bx) \in \SOS{\bx} \label{eq:lam:constr}
    \end{align}
\end{subequations}
The multiplier degree $n_\lambda$ is chosen a priori; starting from $n_\lambda=0$, if the resulting SDP is infeasible we increase $n_\lambda$ and retry, since higher degrees often improve feasibility in practice. Once $\rho$ is optimized, the Lyapunov function is normalized so that the newly identified level set $\Omega$ corresponds to the unit value.

In the subsequent $\mu$-step ({\bf Step 2}), we keep the normalized $V(\bx)$ fixed and find the multiplier $\mu(\bx)$ in constraint~\eqref{eq:SOSROAshape} so as to enlarge the ellipsoidal set $S=\{\bx: \bx^T \bP \bx \le 1\}$ within the previously obtained $\Omega$:
\begin{subequations}\label{eq:mu:step}
    \begin{align}
        \underset{\bP, \mu(\bx)}{\text{minimize}} \quad & \tr(\bP) \label{eq:mu:obj} \\
        \text{s.t.} \quad           & \norm{\bx}^{2d_1} (\bx^T \bP \bx -1) + \norm{\bx}^{2d_2} \mu(\bx) (V(\bx)-1) \in \SOS{\bx} \label{eq:mu:shape}
    \end{align}
\end{subequations}
The multiplier degree $n_\mu$ is chosen a priori, and the same try-and-update logic as in the $\lambda$-step is implemented.

Finally, in the $V$-step ({\bf Step 3}) we solve a Lyapunov-update problem similar to~\eqref{eq:SOSROA}, but with the multipliers $\lambda(\bx)$ and $\mu(\bx)$ fixed to the values computed in the first two steps (thereby eliminating the bilinear terms), to update $V(\bx)$ for the next iteration:
\begin{subequations}\label{eq:V:step}
    \begin{align}
        \underset{V(\bx),\bP}{\text{minimize}} \quad & \tr(\bP) \label{eq:V:cost} \\
        \text{s.t.} \quad           & V(\bx) \in \SOSs{\bx}     \label{eq:V:v}  \\
        \phantom{\text{s.t.} \quad} & V(\bzero) = 0           \label{eq:V:v0}  \\
        \phantom{\text{s.t.} \quad} & \norm{\bx}^{2d} (V(\bx)-1)+\lambda(\bx) \dV(\bx) \in \SOS{\bx} \label{eq:V:in} \\
        \phantom{\text{s.t.} \quad} & \norm{\bx}^{2d_1} (\bx^T \bP \bx -1) + \norm{\bx}^{2d_2} \mu(\bx) (V(\bx)-1) \in \SOS{\bx} \label{eq:V:shape}
    \end{align}
\end{subequations}

The described procedure, applied to the benchmark system~\eqref{eq:vdp}, allows to approximate the origin's ROA very accurately using a sixth-degree Lyapunov function. As shown in Figure~\ref{fig:vdp:ROAiter}, the boundary of the estimated ROA (pink region with red boundary) closely tracks the system's limit cycle (black line), i.e., the boundary of the exact ROA.

\subsection{Safe subset with state constraints}
\label{subsec:safeset}

\begin{figure}
    \centering
     \includegraphics[width=0.48\linewidth]{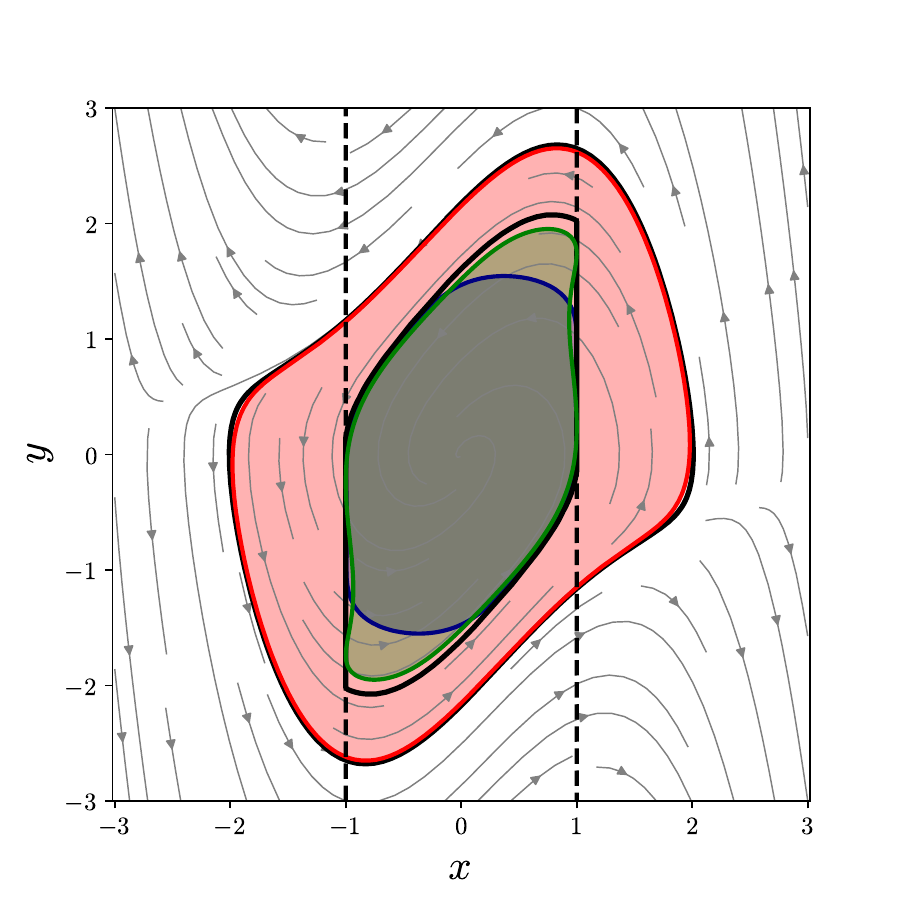}
    \caption{Classical ROA estimate with no constraints (pink region, red boundary). State constraints $x^2\leq 1$ are introduced (dashed black lines). The boundary of the true safe ROA is illustrated in solid black lines. The safe ROA estimate computed enforcing state constraints is given by the blue region (blue boundary). The safe ROA estimate given by the green region (green boundary) is obtained introducing also anchor points as described in the hybrid approach.}
    \label{fig:vdp:safe}
\end{figure}

In many control applications the operating envelope is constrained not only by stability criteria, but also by safety or comfort requirements that bound the admissible trajectories of the system. 
We therefore target, for the Van der Pol oscillator, an invariant subset that satisfies a prescribed collection of state inequalities, defined as safe ROA.
The SOS framework accommodates for state constraints with minor adjustments to the alternating scheme described above.

To illustrate the idea, we enforce the constraint $x^2\le1$ to our benchmark system~\eqref{eq:vdp}, i.e., we require that trajectories originating inside the estimated safe set must remain within the strip delimited by the dashed black lines in Figure~\ref{fig:vdp:safe}.
Adopting again the S-procedure, the state constraint is given by
\begin{equation}
    \label{equ:vdpcon}
    (V(\bx)-\rho)+\eta(\bx) (x^2-1) \in \SOS{x},
\end{equation}
and is added to the $\lambda$- and $V$-steps (Step 1 and Step 3 respectively) of the iterative procedure, i.e., problems~\eqref{eq:lam:step} and~\eqref{eq:V:step}. 
During the $\lambda$-step (Step 1), we introduce and optimize the new polynomial multiplier $\eta(\bx)$ tied to the set inclusion $\Omega\subseteq\{\bx:x^2\le1\}$. In the subsequent $V$-step (Step 3), $\eta(\bx)$ is frozen to the value found in the $\lambda$-step and the same SOS constraint is enforced while updating $V(\bx)$.

The blue region in Figure~\ref{fig:vdp:safe} is the estimated safe ROA resulting from the constrained SOS computation: it is an invariant subset of the unconstrained ROA (red region) that also fulfils the additional state bounds $x^2\leq 1$. 

For comparison, the solid black line tangent to the estimated safe ROA represent the exact safe ROA boundary, obtained via exhaustive simulations starting from sampled initial conditions all over the state space.
Although more conservative with respect to the ROA estimate, the estimated safe ROA still tracks the target region closely and preserves its main geometric features.

Furthermore, we explored a hybrid method to tighten the estimation of the safe ROA. 
While an exhaustive simulation sweep over the state space is often impractical, a small set of sampled boundary points of the safe ROA can efficiently guide the SOS solution to higher levels of accuracy.
The hybrid method requires a preliminary step, during which a few points on the searched safe set are identified. For example, by sampling initial conditions on the set $y=2x$ and computing the resulting trajectories, we identify two target points $\bx_1=(1,2)$ and $\bx_2=(-1,-2)$, laying approximately near the top right and bottom left vertices of the target safe ROA.

We use $\bx_1$ and $\bx_2$ as \emph{anchor points} for the estimated safe ROA by reformulating the $V$-step as
\begin{subequations}\label{eq:safeVstep}
    \begin{align}
        \underset{V(\bx), \bP, \gamma}{\text{minimize}} 
        \quad & w_1 \, \tr \bP + w_2\, \gamma \\
        \text{s.t.} \quad           & V(\bx) \in \SOSs{\bx}       \\
        \phantom{\text{s.t.} \quad} & V(\bzero) = 0            \\ 
        \phantom{\text{s.t.} \quad} & V(\bx_1) = \gamma  \\ 
        \phantom{\text{s.t.} \quad} & V(\bx_2) = \gamma  \\ 
        \phantom{\text{s.t.} \quad} & \norm{\bx}^{2d} (V(\bx)-1)+\lambda(\bx) \dV(\bx) \in \SOS{\bx}  \\
        \phantom{\text{s.t.} \quad} & \norm{\bx}^{2d_1} (\bx^T \bP \bx -1) + \norm{\bx}^{2d_2} \mu(\bx) (V(\bx)-1) \in \SOS{\bx} \\
        \phantom{\text{s.t.} \quad} & (V(\bx)-\rho)+\eta(\bx) (x^2-1) \in \SOS{x} \label{equ:safecon2}
    \end{align}
\end{subequations}
where $\gamma$ is a slack variable representing the symbolic value of the unknown Lyapunov function at the anchor points, and $w_1$ and $w_2$ are weights for the cost function. In our example these are set to $w_1=0.9$ and $w_2=0.1$, respectively.
Intuitively, the boundary of the Lyapunov level set $V(\bx)=1$ is ``pulled'' toward the sampled anchor points lying on the exact target region. As $\gamma$ decreases toward 1, the level set $V(\bx)=1$ progressively tightens around these anchor points. The state constraint~\eqref{equ:safecon2} prevents the boundary of the safe set from extending beyond them. The outcome of the hybrid method is illustrated by the green region in Figure~\ref{fig:vdp:safe}, which provides a more accurate approximation than the basic version shown in blue.

\section{SOS-friendly vehicle-with-driver model}
\label{sec:vehicle}

\subsection{Vehicle-with-driver model}
\label{subsec:model}

The SOS procedure is applied to the vehicle-with-driver model introduced in~\cite{Mastinu:GlobalStabilityRoad:2023}, which is briefly summarized here for completeness. Despite its simplicity, this model already offers quantitative insight into the stability of the coupled driver-vehicle dynamics, as validated in~\cite{Mastinu:HowDriversLose:2024}.

\begin{figure}
    \centering
    \includegraphics[width=0.90\textwidth]{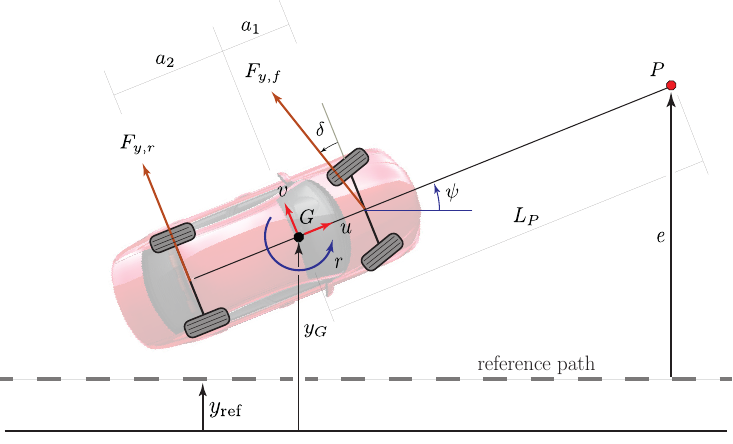}
    \caption{Vehicle-with-driver model. The delayed steering logic mimics a human driver by correcting the lateral error $e$ with respect to the reference path (grey dashed line) of the lookahead point $P$ (red).}
    \label{fig:model}
\end{figure}

The system relies on a single-track model with fixed longitudinal speed $u$ and uses an internal delayed law for the front steering angle $\de$ that drives the vehicle to follow a reference straight path (see Figure~\ref{fig:model}). 
The delayed steering logic mimics a human driver with a time delay of $\tau=\SI{0.2}{s}$ and corrects the lateral error $e$, with respect to the reference path, of a lookahead point $P$. The lookahead distance $\Lprev$ of $P$ from the vehicle centre of mass $G$ is computed as $\Lprev = \frac{u}{\Tprev}$, with the preview time $\Tprev$ fixed at $\SI{0.5}{s}$.

The dynamics of the system is described by the following set of equations:
\begin{subequations}\label{eq:model}
    \begin{align}
        m\, \dv &= - m\, u\, r + \Fyf \cos\de + \Fyr   \label{eq:model:v}\\
        J\, \dr &= a_1 \Fyf \cos\de - a_2 \Fyr     \label{eq:model:r}\\
        \dyG  &= u \sin\psi +v \cos\psi             \label{eq:model:yG}\\
        \dpsi &= r                                  \label{eq:model:psi}\\
        \dde &= \deone                         \label{eq:model:de}\\
        \ddeone &= \detwo                      \label{eq:model:dde}\\
        \ddetwo &= \frac{6}{\tau^3} \left(-\de -\tau\, \ddeone - \frac{\tau^2}{2}\detwo - k\,e - k_d\, \ed\right)                                \label{eq:model:ddde}
    \end{align}
\end{subequations}
The descriptions of the variables and parameters appearing in~\eqref{eq:model} are summarized in Table~\ref{tab:model}.
\begin{table}
    \centering
    \caption{Variables and parameters of the vehicle-with-driver model in~\eqref{eq:model}.}
    \label{tab:model}
    \begin{tabular}{clcl}
        \toprule
        Symbol & Description & Symbol & Description \\
        \midrule
        $m$      & vehicle mass                 & $\de$      & front axle steering angle \\
        $J$      & yaw moment of inertia        & $\deone$   & front axle steering rate\\
        $a_1$    & distance of $G$ from the front axle & $\detwo$   & front axle steering angular acceleration \\
        $a_2$    & distance of $G$ from the rear axle & $\Fyf$     & lateral force acting on the front axle \\
        $u$      & longitudinal speed           & $\Fyr$     & lateral force acting on the rear axle \\
        $v$      & lateral speed                & $e$        & lateral error of preview point $P$ \\   
        $r$      & yaw rate                     & $\tau$     & steering delay time\\
        $\yG$    & lateral displacement of $G$ & $k$        & proportional steering gain \\
        $\psi$   & yaw angle                    & $k_d$      & derivative steering gain \\
        \bottomrule
    \end{tabular}
\end{table}
Equations~\eqref{eq:model:de}--\eqref{eq:model:ddde} implement a delayed PD steering law on the lateral error $e=\yG-y_{\text{ref}}+\Lprev\sin\psi$, with proportional gain $k$ and derivative gain $k_d$:
\begin{equation}
    \label{eq:pd}
    \de(t+\tau)=-k\,e-k_d\,\ed \,.
\end{equation}
The left hand term of~\eqref{eq:pd} is approximated through the third-order Taylor expansion
\begin{equation}
    \label{eq:taylor}
    \de(t+\tau)\approx\de(t)+\dde(t)\,\tau+\frac{\ddde(t)}{2}\,\tau^2+\frac{\dddde(t)}{6}\,\tau^3.
\end{equation}
which yields~\eqref{eq:model:de}--\eqref{eq:model:ddde}. Deeper insights on the accuracy of the adopted approximation can be found in~\cite{Mastinu:GlobalStabilityRoad:2023}.

The described vehicle-with-driver system has a fixed point at the origin, i.e., for $\bx=\left[v\;r\;\yG\;\psi\;\de\;\deone\; \detwo\right]^T=\bzero$. 
The stability of this equilibrium depends on the system's parameters; for realistic settings one can usually identify a critical longitudinal speed below which the origin is asymptotically stable~\cite{Mastinu:GlobalStabilityRoad:2023}. 
In what follows, we focus on such operating conditions and estimate a safe invariant subset of the ROA of the stable equilibrium at the origin.

\subsection{Polynomial approximation}
\label{subsec:polynomial}
The SOS procedure for ROA estimation introduced in Sec.~\ref{sec:roa} presumes polynomial system dynamics, a condition that the model in~\eqref{eq:model} clearly violates.
Two sources break the polynomial structure: the trigonometric terms in $\psi$ and $\de$ (noting that $e=\yG-y_{\text{ref}}+\Lprev\sin\psi$ carries one as well), and the tire constitutive law,
which we model through a standard Magic Formula (MF)
\begin{equation}
    \label{equ:magic}
    \Phii(\ali) = D_i \sin\left\{ C_i \arctan \left[B_i \, \ali - E_i \,\big(B_i \,\ali - \arctan(B_i \,\ali)\big) \right] \right\}, \quad i=\{f,r\}
\end{equation}
mapping the axle slip angles $\alpha_f$ and $\alpha_r$ to the lateral front and rear forces $\Fyf$ and $\Fyr$, respectively. The axle slip angles depend linearly on the state variables ($u$ is constant),
\begin{equation}
    \label{eq:congruence}
    \alf = \de - \frac{v + a_1\, r}{u} , 
    \qquad
    \alr = - \frac{v - a_2\, r}{u},
\end{equation}
yet the Magic-Formula relationships $\Fyf = \Phif(\alf)$ and $\Fyr = \Phir(\alr)$ remain non-polynomial, which prevents a direct SOS treatment.
Including longitudinal dynamics would introduce an additional non-polynomial term in the congruence equations in~\eqref{eq:congruence}, since the denominator $u$ would become a state rather than a parameter.

A polynomial approximation of model~\eqref{eq:model} is thus required, noting that higher polynomial degrees entail increased computational cost in the SOS procedure. We proceed in two steps.
First, trigonometric terms in $\de$ and $\psi$ are replaced by Taylor expansions. Given that $\de \in [\ang{-5},\ang{5}]$, we set $\cos\de \approx 1$. Larger yaw excursions motivate third-order approximations, 
$\sin\psi \approx \psi - \psi^3/6$ and 
$\cos\psi \approx 1 - \psi^2/2$, which remain accurate for 
$\psi \in [\ang{-50},\ang{50}]$. 
Second, the axle characteristics are approximated using cubic fits of 
$\Phif(\alf)$ and $\Phir(\alr)$.

\begin{figure}[ht]
    \centering
    \includegraphics[width=0.6\textwidth]{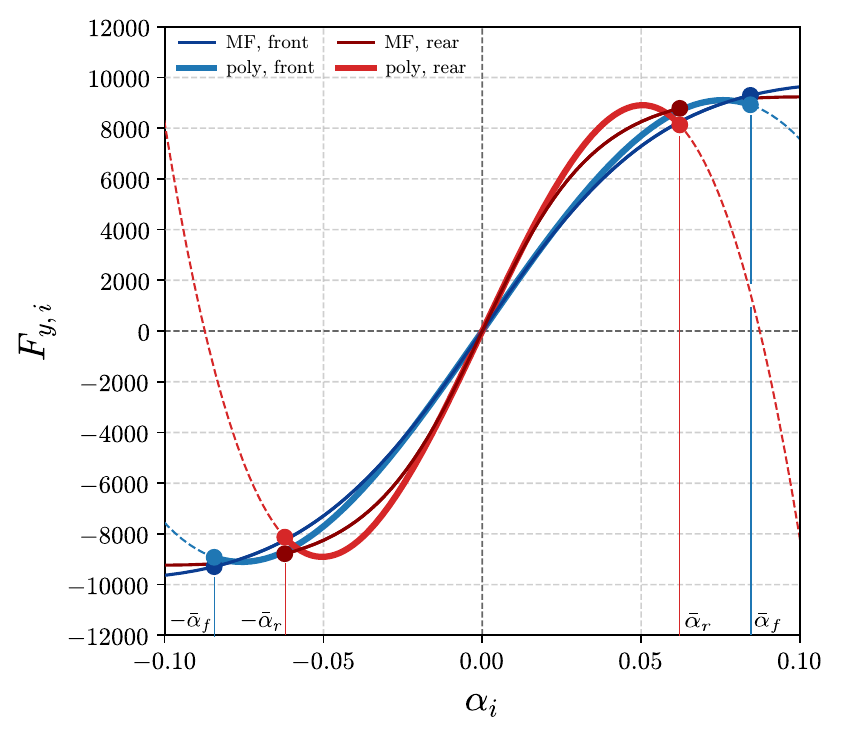}
    \caption{Polynomial fitting of axle characteristics in the understeering setup. Dark blue/red thin curves show the Magic Formula (MF) for the front/rear axles ($\alpha_i$ in rad, $F_{y,i}$ in N), while the light counterparts depict the cubic fits (poly), solid within the $[-\alilim,\alilim]$ range used for fitting and dashed outside it. Vertical blue/red lines mark the axle slip angles limits corresponding to $95\%$ of the respective Magic Formula peak forces.}
    \label{fig:axle}
\end{figure}

Because polynomials cannot reproduce the saturation inherent in the 
Magic Formula, the fitting domain is limited to 95\% of the peak lateral 
force, yielding slip-angle thresholds $\alflim$ and $\alrlim$.
Cubic fits are then computed over 
$\alf \in [-\alflim,\alflim]$ and 
$\alr \in [-\alrlim,\alrlim]$, as illustrated in 
Figure~\ref{fig:axle}.
The resulting polynomials provide an accurate representation only within these bands. 
Consequently, when computing constrained invariant subsets of the ROA we enforce as validity constraints
\begin{equation}
    \label{equ:alpha_con}
    |\alf(\bx(t))| \le \alflim, \qquad |\alr(\bx(t))| \le \alrlim, 
    \qquad \forall t \ge0.
\end{equation}
This ensures that the certified ROA remains entirely within the region where the polynomial approximation of the axle characteristics is reliable. 

No additional bounds are imposed on the steering angle $\de$ or the yaw angle $\psi$, as the invariant subsets of the ROA remained within their validity limits in all examined cases. 
If needed under different operating conditions, constraints analogous to those on the axle slip angles could be introduced. With the above substitutions, the polynomial approximation of the vehicle-with-driver model is of third degree.

\subsection{SOS setup for the vehicle-with-driver}
The vehicle-with-driver model has seven states. 
As recalled in Sec.~\ref{subsec:SOSform}, applying an SOS constraint to a polynomial requires expressing the polynomial through its Gram matrix, whose dimension grows rapidly with the polynomial degree and the number of variables.
Gram-matrix entries are a decision variables in the SOS-based ROA algorithm of Sec.~\ref{subsec:iterSOS}, so keeping a low degree for the Lyapunov polynomial---we fix it at four---is a practical compromise that avoids a prohibitive number of decision variables.

The symmetry of the vehicle-with-driver model is exploited by restricting $V(\bx)$ and the multipliers $\lambda(\bx)$ and $\mu(\bx)$ introduced in Sec.~\ref{subsec:iterSOS} to \emph{even} polynomials. 
This is achieved by constraining to zero the coefficients of the odd terms of $\lambda(\bx)$ in the $\lambda$-step ({\bf Step 1}), of $\mu(\bx)$ in the $\mu$-step ({\bf Step 2}), and of $V(\bx)$ in the $V$-step ({\bf Step 3}).
The simmetry constraints do not reduce the number of variables, but help directing the solution toward even polynomials, matching the physics of the vehicle-with-driver model. 
The exponents $d$, $d_1$, and $d_2$ introduced in~\eqref{eq:SOSROAin} and~\eqref{eq:SOSROAshape} are set to $2$, $1$, and $0$, respectively.

The state constraints~\eqref{equ:alpha_con} on the axle slip angles are enforced via the S-procedure, i.e.,
\begin{equation}
    \label{equ:alphaSOS}
    (V(\bx)-\rho) + \etai(\bx)(\ali(\bx)-\alilim) \in \SOS{\bx}, \qquad i\in\{f,r\}.
\end{equation}
where $\etai(\bx)$ are polynomial multipliers of degree $\deg(V)-1$, i.e., third degree. 
The symmetry of $V(\bx)$ guarantees that also the lower boundary $\ali(\bx)\ge-\alilim$ is fulfilled.
Constraints such as~\eqref{equ:alphaSOS} are enforced in both the 
$\lambda$-step ({\bf Step 1}) and the $V$-step ({\bf Step 3}). 
In {\bf Step 1}, the multipliers $\eta_i(\bx)$ are optimized jointly with 
$\lambda(\bx)$. In {\bf Step 3}, the same SOS constraints are imposed while 
updating $V(\bx)$, but the multipliers $\eta_i(\bx)$ are kept fixed at the 
values obtained in {\bf Step 1}, mirroring the treatment of $\lambda(\bx)$.

\section{Vehicle-with-driver safe invariant set}
\label{sec:results}
\subsection{Tested scenarios}
Results of the proposed procedure on the vehicle-with-driver model described in Sec.~\ref{sec:vehicle} are presented for two different scenarios: an oversteering (OV) and an understeering (UN) vehicle. 
In both cases the longitudinal speed is fixed at $u=\SI{90}{km/h}$, while the axle characteristics $\Phif(\alf)$ and $\Phir(\alr)$ are described by the corresponding Magic Formula parameters listed in Table~\ref{tab:scenarios}. 
The controller gains $k$ and $k_d$ also differ between scenarios, with values listed in Table~\ref{tab:scenarios}. 
These gains were obtained by fitting experimental data~\cite{Mastinu:GlobalStabilityRoad:2023} and depend on both driver behavior and vehicle parameters. 
%
%


\begin{table}
    \centering
    \caption{Parameters of the oversteering (OV) and understeering (UN) scenarios.}
    \label{tab:scenarios}
    \begin{tabular}{ccccccccccc}
        \toprule
        Scenario & $B_f$ & $C_f$ & $D_f$ & $E_f$ & $B_r$ & $C_r$ & $D_r$ & $E_r$ & $k$ & $k_d$ \\
        \midrule
        OV & $14.50$ & $1.89$ & $9778$ & $0.29$ & $13.50$ & $1.45$ & $9234$ & $0.31$ & $0.025$ & $0.004$ \\
        UN & $9.86$ & $1.87$ & $9778$ & $0.28$ & $18.75$ & $1.53$ & $9234$ & $0.30$ & $0.010$ & $0.008$ \\
        \bottomrule
    \end{tabular}
\end{table}

For each scenario, a safe invariant subset of the ROA (SOS-S-ROA) of the stable equilibrium at the origin is computed using the three-step SOS procedure described in Sec.~\ref{subsec:iterSOS}. 
In all cases, this safe set is represented as the region enclosed by the unit level set $V(\bx)=1$ of a degree-4 polynomial Lyapunov function---\emph{this is the final outcome of the procedure}.
\begin{figure}
    \centering
    \subfigure[Intersections with $v$-$r$ plane]{%
        \label{fig:OVvr}%
        \includegraphics[width=0.48\textwidth]{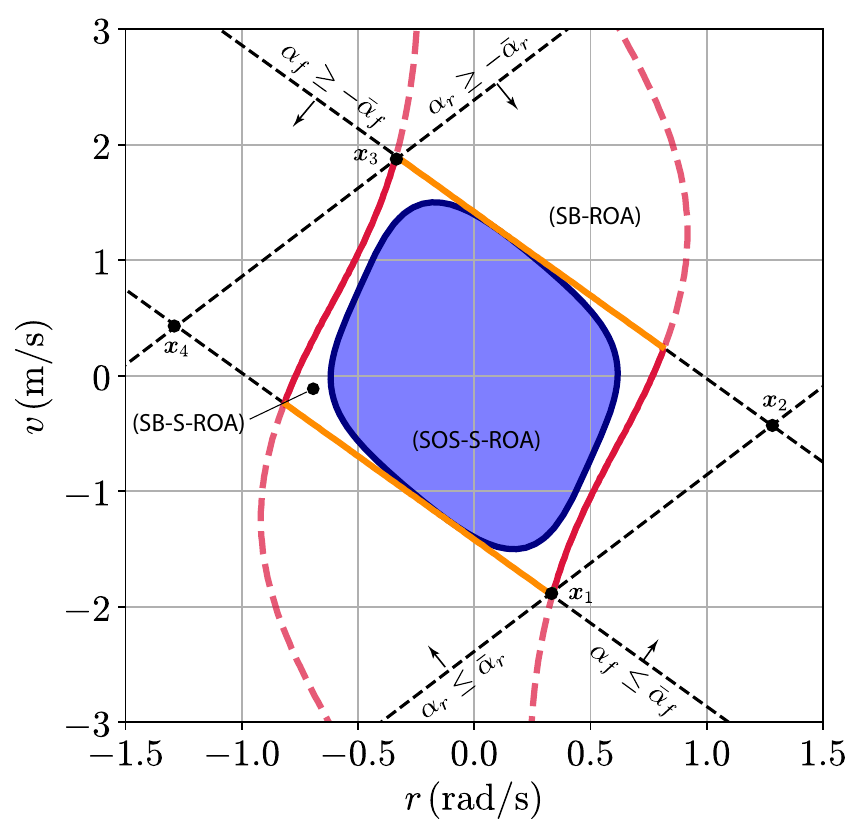}}
    \hfill
    \subfigure[Intersections with $y_G$-$\psi$ plane]{%
        \label{fig:OVypsi}%
        \includegraphics[width=0.48\textwidth]{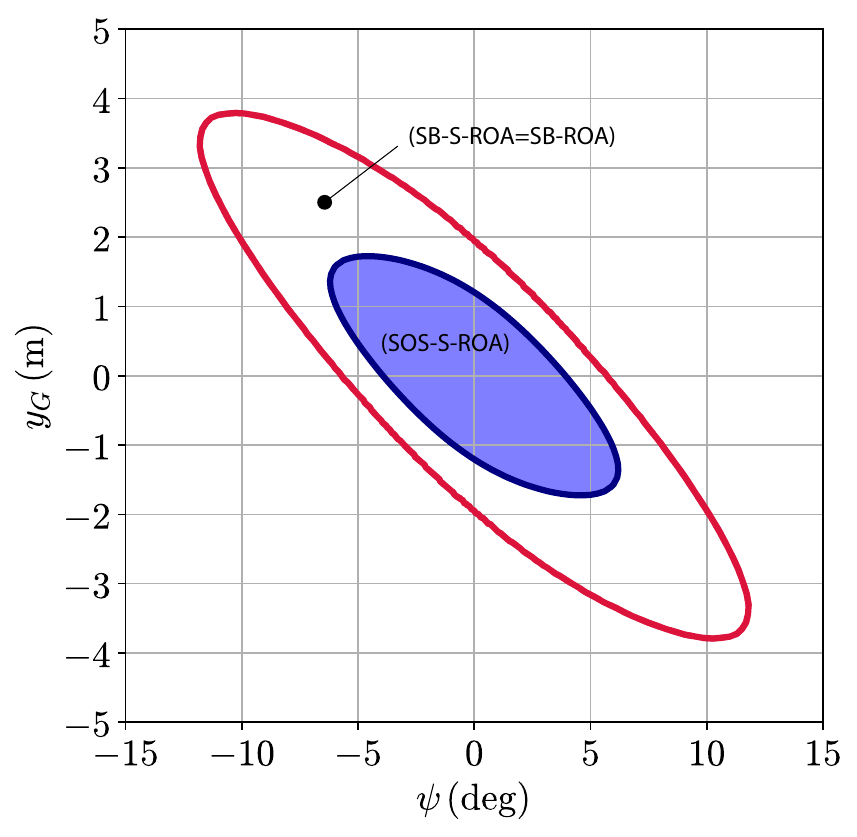}}
    \caption{Oversteering scenario: safe subset of the ROA estimated via SOS, SOS-S-ROA (blue region), state constraints on axle slip angles (dashed black lines), simulation-based ROA, SB-ROA, boundaries (dashed red lines) and safe subset of the simulation-based ROA, SB-S-ROA, boundaries (red for stability limit, yellow for $\alf$ limit) }
    \label{fig:OVplanes}
\end{figure}

\begin{figure}
    \centering
    \subfigure[Intersections with $v$-$r$ plane]{%
        \label{fig:UNvr}%
        \includegraphics[width=0.44\textwidth]{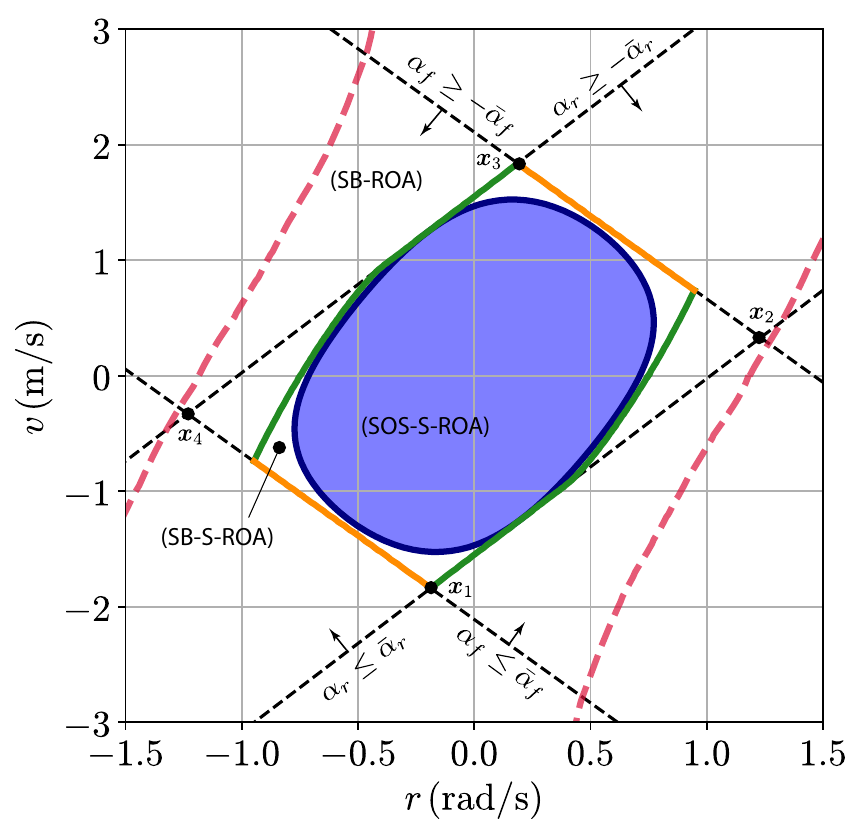}}
    \hfill
    \subfigure[Intersections with $y_G$-$\psi$ plane]{%
        \label{fig:UNypsi}%
        \includegraphics[width=0.5\textwidth]{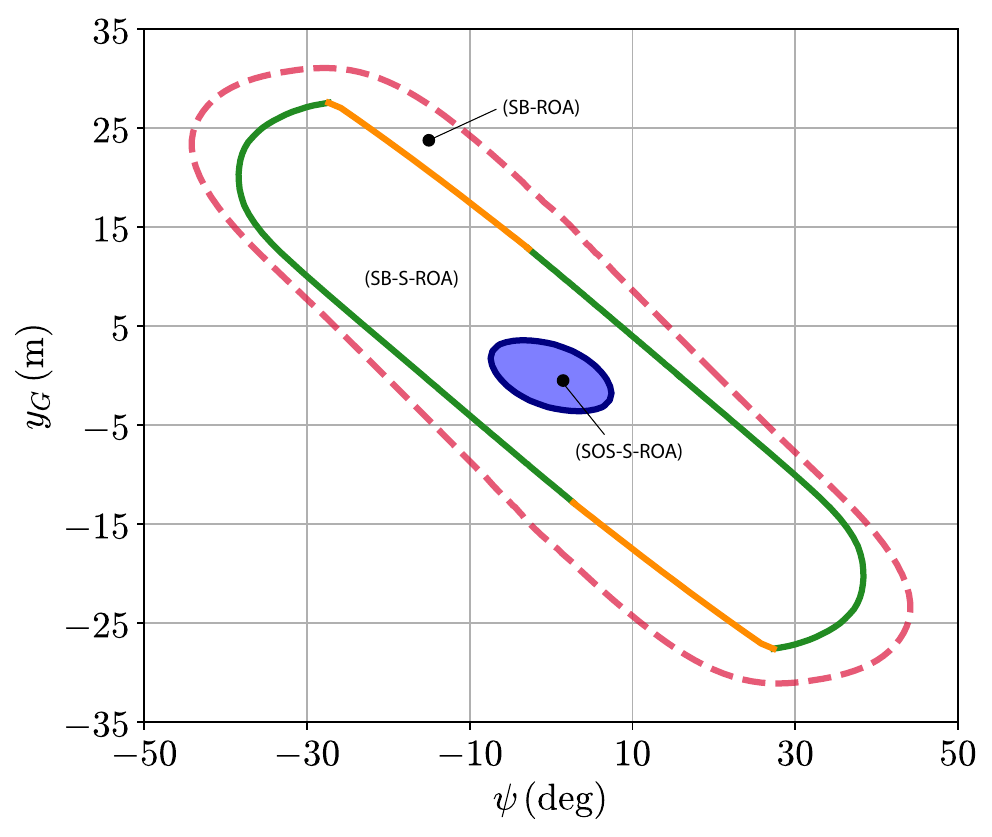}}
    \caption{Understeering scenario: safe subset of the ROA estimated via SOS, SOS-S-ROA (blue region), state constraints on axle slip angles (dashed black lines), simulation-based ROA, SB-ROA, boundaries (dashed red lines) and safe subset of the simulation-based ROA, SB-S-ROA, boundaries (red for stability limit, yellow for $\alf$ limit, green for $\alr$ limit) }
    \label{fig:UNplanes}
\end{figure}
We show the SOS-S-ROAs obtained for both OV and UN vehicles through their \emph{intersections} with two specific planes of the state space: the $v$-$r$ plane (Figures~\ref{fig:OVvr} and~\ref{fig:UNvr}) and the $y_G$-$\psi$ plane (Figures~\ref{fig:OVypsi} and~\ref{fig:UNypsi}). 
On each panel, the blue region represents the intersection of the estimated SOS-S-ROA $\Omega=\{\bx:V(\bx)\le 1\}$ with the corresponding plane.
 
On the $v$-$r$ plane, the constraints \eqref{equ:alpha_con} on axle slip angles are shown to form a parallelogram (dashed black lines), where the bottom left side corresponds to $\alf\leq \bar{\al}_f$, while the bottom right side corresponds to $\alr\leq \bar{\al}_r$. The upper sides correspond to $\alf\geq -\bar{\alpha}_f$ and $\alr \geq -\bar{\alpha}_r$, respectively, and are parallel to their counterparts.
In both OV and UN cases we use the vertices of the parallelogram $\bx_j$ (Figures~\ref{fig:OVvr} and Figures~\ref{fig:UNvr}) in the cost functions of the $V$-step (Step 3) as anchor points for the SOS-S-ROA boundary, following the hybrid method described in~\eqref{eq:safeVstep} Sec.~\ref{subsec:safeset}.
Although these vertices do not necessarily lie on exact safe ROA, they are profitably used since readily available by pairwise intersections of the four slip-angle constraints. 
It is worth noting their role as anchor points: these vertices act as 
\emph{pegs} to which the Lyapunov level set is attached and \emph{stretched} in 
problem~\eqref{eq:safeVstep}.

\subsection{Results}
\label{subsec:res}
To interpret the SOS-S-ROA intersections, consider any initial condition 
$\bx_0=(v_0,r_0,0,0,0,0,0)$ on the $v$--$r$ plane, with $(v_0,r_0)$ lying in 
the blue region of Figure~\ref{fig:OVvr}. Such conditions may arise when the 
vehicle, initially at $\bx=\bzero$, experiences lateral and yaw disturbances 
that primarily excite the $v$--$r$ dynamics.

Membership of $\bx_0$ in the SOS-S-ROA can be immediately verified by checking $V(\bx_0)\le1$, where $V(\bx)$ is the Lyapunov function obtained from the SOS procedure in Algorithm~\eqref{alg:sos_iter}. 
This \emph{certifies} i) convergence of the trajectory $\bx(t)$ originated in $\bx_0$ to the origin and ii) satisfaction of the slip-angle constraints~\eqref{equ:alpha_con} for all $t\ge0$, ensuring the validity of the polynomial approximation. 
It is worth remarking that, although the initial condition lies in the $v$--$r$ plane, the trajectory $\bx(t)$ evolves in the full seven-dimensional state space; nevertheless, its evolution remains confined within the certified SOS-S-ROA manifold (see Figure~\ref{fig:sim}).

A similar interpretation applies to the $y_G$--$\psi$ plane 
(Fig.~\ref{fig:OVypsi}), where $\bx_0=(0,0,y_{G0},\psi_0,0,0,0)$ represents a lateral displacement $y_{G0}$ and an initial heading $\psi_0$ relative to the reference path. Such conditions may arise, for instance, from an abrupt lane-change manoeuvre in which the vehicle is laterally offset and exhibits a nonzero yaw angle at the onset.

%

For comparison, for the full nonlinear vehicle dynamics (with no polynomial approximations), the ROA was computed via a simulation-based approach (SB-ROA). Initial conditions were sampled on each plane and the system was integrated over a \SI{10}{s} horizon. A sampled point was classified as belonging to the simulation-based safe ROA, SB-S-ROA, if the corresponding trajectory converged to the equilibrium and satisfied all state constraints---specifically, the front and rear axle slip-angle bounds---for the entire time interval.

The resulting simulation-based boundaries of both the SB-ROA and the SB-S-ROAs are reported in 
Figures.~\ref{fig:OVvr} and~\ref{fig:OVypsi} for the OV vehicle. They also indicate which constraint is violated immediately outside the SB-S-ROA: loss of asymptotic stability (solid red), violation of the front slip-angle limit (yellow), or violation of the rear slip-angle limit (green). Along the solid red curves (see Figure~\ref{fig:OVvr}), the boundaries of the SB-ROA and of the SB-S-ROA coincide; the portions of the simulation-based ROA where 
slip-angle constraints are not enforced (simply SB-ROA) are shown as dashed red lines. The same observations apply to the UN vehicle in Figure~\ref{fig:UNplanes}.

It is worth remarking that only the intersections of SB-ROA and SB-S-ROA with the $v$-$r$ and $y_G$-$\psi$ planes were computed, since exhaustive sampling of the full seven-dimensional state space is prohibitively expensive. 
By contrast, the SOS-S-ROA is obtained as a seven-dimensional manifold, and the planar sections shown in Figures~\ref{fig:OVplanes} and~\ref{fig:UNplanes} are extracted, through intersections, from this \emph{global} certificate.

The SOS procedure was implemented in Python and executed through the 
\textsc{Drake}~\cite{drake:MIT:2025} optimization suite, employing 
\textsc{Mosek}~\cite{mosek:2025} as the back-end SDP solver.
It was run on a laptop equipped with an Intel i7 processor and \SI{26}{GB} of RAM.
The OV case required \SI{301}{s} and six iterations, the UN case \SI{287}{s} and also six iterations. 
Each step of Algorithm~\eqref{alg:sos_iter} corresponds to a distinct problem with its own set of decision variables: the initialization step has \SI{1296} variables, the $\lambda$-step (Step 1) has \SI{55894} variables, the $\mu$-step (Step 2) has \SI{723} variables, and the $V$-step (Step 3) has \SI{57007} variables.

\subsection{Discussion}
For each panel in Figures~\ref{fig:OVplanes} and~\ref{fig:UNplanes} we address two aspects: i) how the polynomial approximation of the vehicle-with-driver model limits the SB-S-ROA with respect to the SB-ROA, ii) how the SOS-S-ROA tightens to the SB-S-ROA, assumed as exact target estimate.

In the OV case (Figure~\ref{fig:OVplanes}), on the $v$-$r$ plane the SB-S-ROA shares boundaries with the SB-ROA (solid red lines) and with the front slip angle constraints (yellow lines).
This indicates that the full nonlinear system can be unstable without violating the rear slip angle limits (dotted black lines from $x_1$ to $x_2$ and from $x_3$ to $x_4$), e.g., starting from an initial condition lying outside the SB-S-ROA but inside the dotted black parallelogram. 
On the other hand, the front slip angle limits (yellow lines) remove from the SB-S-ROA the upper and lower portions of the SB-ROA: the approximation of the front axle characteristic is therefore more critical, and extending its fitting range beyond axle saturation would result in a larger SB-S-ROA.
The SOS-S-ROA aligns closely to the SB-S-ROA, except at its vertices.

On the $y_G$-$\psi$ plane, the SB-S-ROA coincides exactly with the SB-ROA (or more precisely, with the simply connected portion of the SB-ROA containing the origin~\footnote{A more detailed analysis (omitted here) shows that, on this plane, the 
section of the SB-ROA forms a non-simply-connected set with islands. 
Only its simply-connected component containing the origin coincides with 
the SB-S-ROA.
}).
Here the SOS-S-ROA is visibly less tight than the exact estimate SB-S-ROA. 
The actual safe sets live in $\mathbb{R}^7$ while we display only planar sections, so the \emph{sharpness} of the SB-S-ROA boundary may not be fully captured in these intersections. 
In addition, a fourth-degree Lyapunov polynomial lacks the flexibility to reproduce such sharp features. 
Therefore, in our view, this loss of accuracy stems from the polynomial-degree constraint required to keep the SOS problem tractable in the present formulation.

For the UN case (Figure~\ref{fig:UNplanes}), analogous 
considerations apply. 
On the $v$-$r$ plane (Figure~\ref{fig:UNvr}) the influence of 
the axle slip angle bounds---which are needed to preserve the 
validity of the polynomial approximation---is even more 
evident. 
Here, the SB-S-ROA lies strictly inside the SB-ROA, so instability of the full nonlinear system only arises if the axle slip angle constraints are breached. 
The discrepancies of the SB-S-ROA boundaries from the dashed black parallelogram at higher $|r|$ highlight their intrinsic difference: 
an initial configuration lying inside the parallelogram, yet outside the SB-S-ROA, satisfies the slip-angle constraints at the initial instant, but violates them along the trajectory---outside the $v$-$r$ plane.

On the $y_G$-$\psi$ plane (Figure~\ref{fig:UNypsi}), we first point out the different axis scales with respect to Figure~\ref{fig:OVypsi}: both the SB-ROA and SB-S-ROA extend farther, which is consistent with the greater stability typically shown by understeering vehicles with respect to oversteering ones.
The SB-S-ROA remains strictly contained within the SB-ROA, with the  
axle slip angle limits preventing instability. 
The SOS-S-ROA, however, does not adhere as tightly to the broader SB-S-ROA. 
This gap is mainly attributable to the limited flexibility of the fourth-degree Lyapunov polynomial and to the sharp features of the SB-S-ROA boundary outside the displayed slice. 
Despite these limitations, the certified region still spans a 
practically relevant operating envelope for the 
understeering scenario.

\begin{figure}
    \centering
    \subfigure[Trajectories' projections on $v$-$r$-$\de$ subspace. The boundaries of intersections of SB-ROA (thin black line), SB-S-ROA (thick black line) and SOS-S-ROA (blue line) with the $v$-$r$ plane are reported.]{%
        \label{fig:sim:vrd}%
        \includegraphics[width=0.49\textwidth]{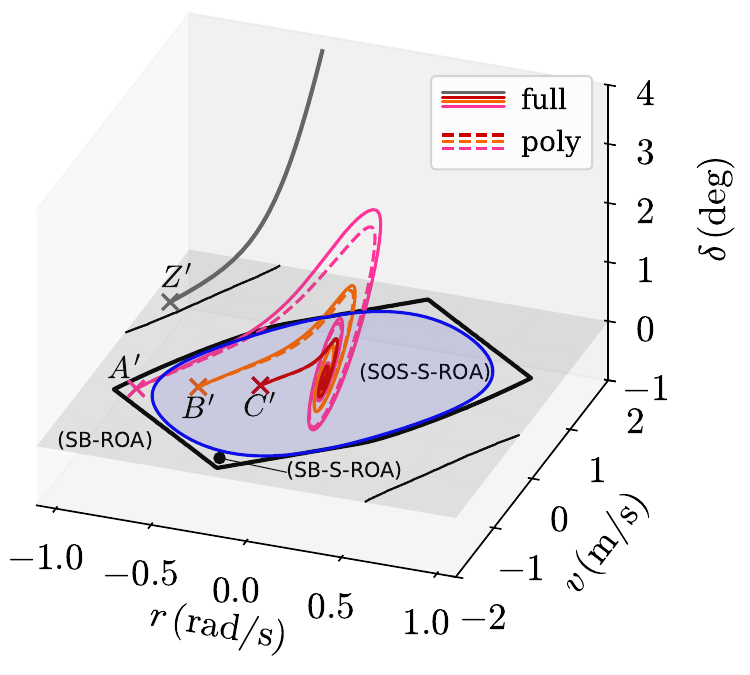}}
    \hfill
    \subfigure[Trajectories' projections on $y_G$-$\psi$ plane.]{%
        \label{fig:sim:ypsi}%
        \includegraphics[width=0.49\textwidth]{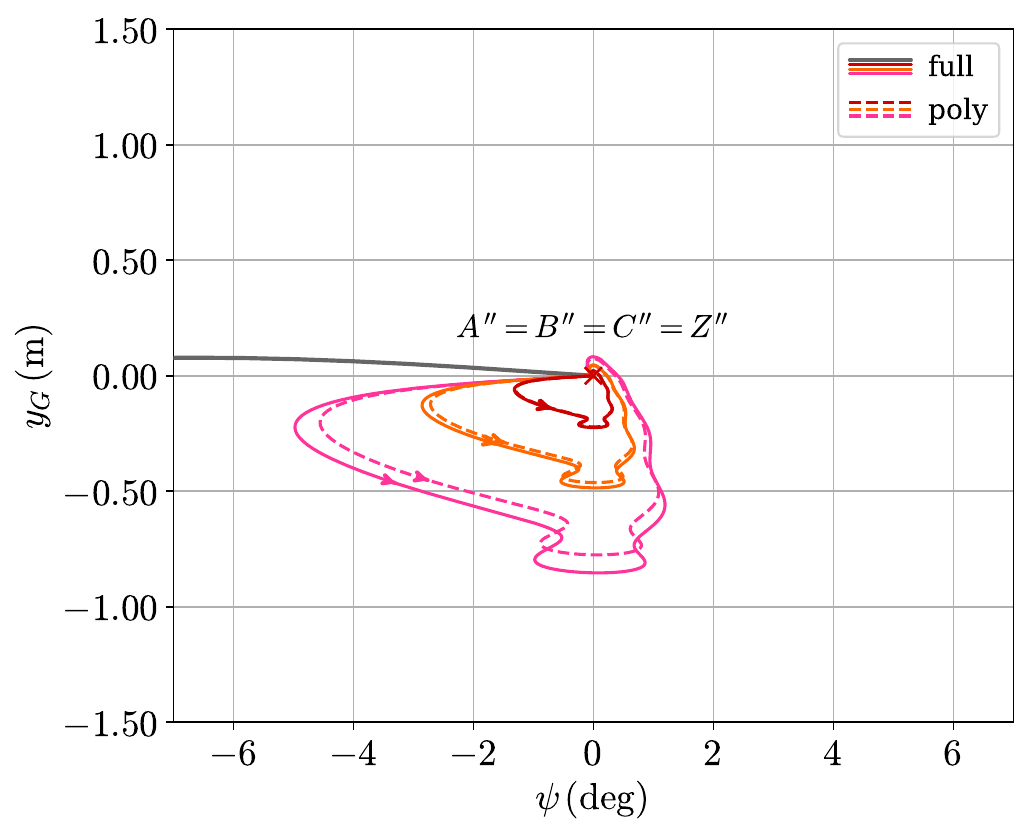}}
    \hfill
    \subfigure[Trajectories' projections on $\alf$-$\alr$ plane, with axle slip angle limits (dashed black lines). ]{%
        \label{fig:sim:alpha}%
        \includegraphics[width=0.49\textwidth]{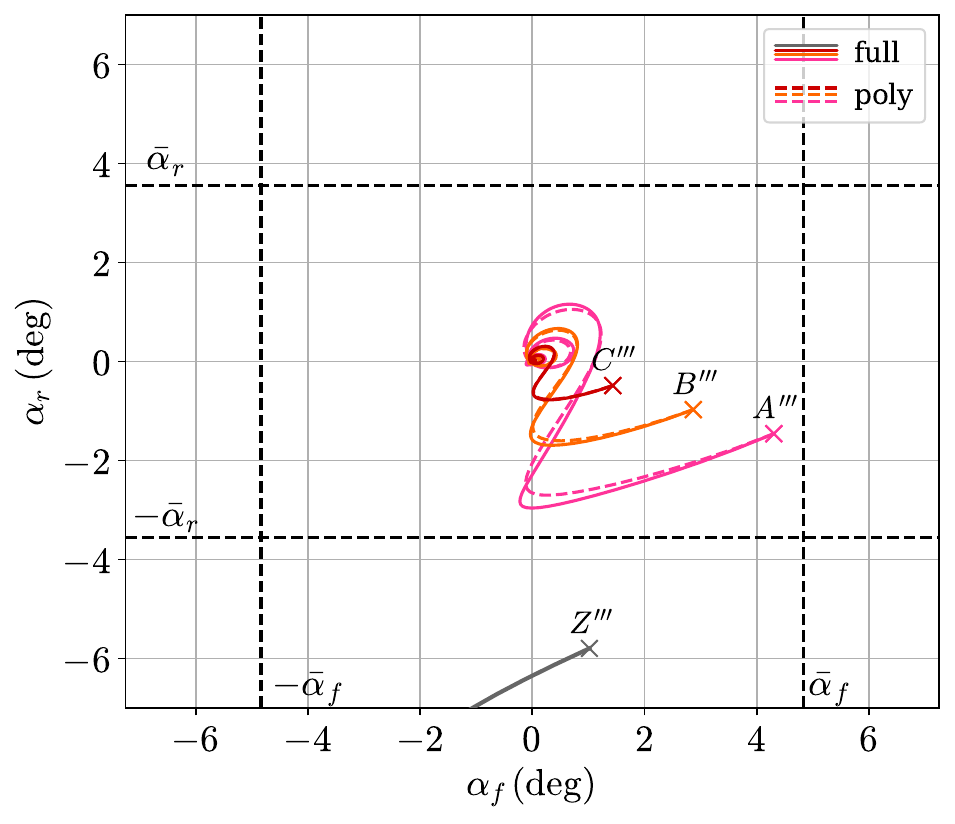}}
    \caption{Trajectories for the understeering case, starting from initial conditions A, B, C, and Z sampled on the $v$-$r$ plane. Trajectories relative to full nonlinear system (solid pink, orange, red and grey lines) and its polynomial approximation (dashed pink, orange and red) are shown.}
    \label{fig:sim}
\end{figure}

Figure~\ref{fig:sim} shows trajectories obtained via simulation for the full nonlinear system (full, solid lines) and the polynomial approximated dynamics (poly, dashed lines) in the UN case. 
The trajectories are projected on the $v$-$r$-$\de$ subspace (panel~\ref{fig:sim:vrd}), the $y_G$-$\psi$ plane (panel~\ref{fig:sim:ypsi}), and the $\alr$-$\alf$ plane (panel~\ref{fig:sim:alpha}).
On the $v$-$r$-$\de$ subspace (panel~\ref{fig:sim:vrd}) the boundaries of intersections of SB-ROA (thin black line), SB-S-ROA (thick black line) and SOS-S-ROA (blue line) with the $v$-$r$ plane are reported.
On the $\alr$-$\alf$ plane (panel~\ref{fig:sim:alpha}) the axle slip angle limits are shown in dashed black lines. 
Four initial conditions are sampled on the $v$-$r$ plane. Each one, e.g. $A$, is shown in each panel through its projected points $A'$,$A''$, and $A'''$. Projected points of configurations $B$,$C$, and $Z$ follow the same nomenclature. As shown on panel~\ref{fig:sim:vrd}, configuration $A$ lies inside the SB-S-ROA but outside the SOS-S-ROA, $B$ and $C$ lie inside the SOS-S-ROA, $Z$ lies outside the SB-ROA.

For initial configuration $Z$, since it lies outside the range of validity of the polynomial approximation, only the diverging full nonlinear trajectory is displayed.
Conversely, for initial configurations $A$,$B$, and $C$ we integrate both the full nonlinear dynamics (solid curves) and the polynomial dynamics (dashed curves).
For each pair of trajectories originating from the same configuration, e.g., the pink trajectories starting from $A$, it can be observed that the evolution of the polynomial dynamics closely tracks the full nonlinear trajectory on all projections, corroborating the validity of the polynomial approximation introduced in Sec.~\ref{subsec:polynomial}.

Since $A'$ lies near a vertex of the SB-S-ROA in the $v$--$r$ plane (panel~\ref{fig:sim:vrd}), its proximity to both front and rear slip limits is clearly reflected in panel~\ref{fig:sim:alpha}. The projection $A'''$ starts close to the upper front slip-angle bound $\alflim$, and the pink trajectories originating from $A'''$ bend downward toward the rear slip-angle lower limit $-\alrlim$. This shows that a vertex of the SB-S-ROA corresponds to simultaneous closeness to both axle slip constraints.  
If point $A$ were placed exactly at the SB-S-ROA vertex, the lower knee of the pink curve (full, solid line) in panel~\ref{fig:sim:alpha} would become tangent to the rear slip-angle lower limit $\alr=-\alrlim$.
In addition, panel~\ref{fig:sim:ypsi} highlights that the initial perturbation, although applied on states not visible in the $y_G$--$\psi$ plane, still leads back to stability after a single orbit, further confirming the consistency of the SB-S-ROA prediction.

Summarizing the findings, in each case the resulting Lyapunov polynomial $V(\bx)$ provides a global certificate of a safe invariant subset of the ROA in the full seven-dimensional state space. Once $V(\bx)$ is computed, assessing the safety of the vehicle-with-driver system reduces to a single inequality check, namely verifying whether $V(\bx)\le 1$ at the current state. This stands in stark contrast to the SB-S-ROA approach, used here only as a validation tool, which is impractical both because exhaustively sampling all possible initial conditions is infeasible and because determining whether a given state belongs to the sampled ROA is not reliably decidable.

If the human driver is well represented by the controller structure adopted in the model, the availability of such a certified SOS-S-ROA enables a fast and reliable stability assessment: evaluating $V(\bx)$ immediately determines whether the closed-loop driver-vehicle system will return to the equilibrium or is instead headed toward instability. This capability could be used as a supervisory trigger for active safety controllers that override or modulate the driver's inputs when the system state approaches the boundary of the safe ROA. 
The same methodology could also be applied to characterize the SOS-S-ROA of a 
vehicle with ESP system by adjusting the time-delay of the model and the 
control logic accordingly.

\section{Conclusions}
\label{sec:conclusion}

This work presented a preliminary application of SOS programming to estimate 
safe invariant sets for a vehicle-with-driver system. The proposed approach 
extends Lyapunov-based ROA estimation to a seven-dimensional coupled 
driver--vehicle model, explicitly incorporating the driver's action into the 
stability assessment.

The results show that SOS techniques can provide certified safety regions for 
the vehicle-with-driver system through the Lyapunov polynomial $V(\bx)$, 
allowing efficient online safety evaluation via a single function check. 
Comparison with simulation-based boundaries indicates that, despite their 
inherent conservatism, the SOS-based estimates capture the main features of the prescribed safety sets and offer rigorous guarantees on both stability and state-constraint satisfaction.

Several limitations warrant further investigation. The polynomial approximation required for SOS tractability imposes restrictive slip-angle bounds, which significantly affect the estimated safe region, particularly in the understeering case. In addition, the simplified single-track model and basic driver representation reduce applicability to more complex operating conditions. 
These approximations are necessary to control computational complexity, but they limit the domain in which the polynomial dynamics remain accurate.

Despite these constraints, the results are promising and demonstrate the 
feasibility of the approach. The methodology provides a systematic framework for certifying safety that can be extended to richer vehicle models and more 
realistic driver behaviors. Future work should investigate higher-degree 
polynomial approximations, improved representations of tire forces, and 
applications to more complex maneuvers beyond straight-line operation. 
A particularly promising direction is the integration of such certified safe regions as supervisory layers in active control systems, potentially contributing to enhance active safety in passenger cars.


\section*{Acknowledgement}
The authors would like to thank Prof. Tobia Marcucci from UCSB for his early guidance, which provided valuable insights into Sum-of-Squares techniques and helped shape the initial direction of this work.


\section*{Disclosure statement}
No potential conflict of interest was reported by the author(s).

\section*{Funding}
This work is supported by project PRIN 2022 PNRR ``Global Stability of road vehicle
motion - STAVE'' CUP I53D23005670001.


\bibliography{references}

\end{document}